\documentclass[]{interact}

\usepackage{epstopdf}
\usepackage[caption=false]{subfig}
\usepackage{cite}
\usepackage{amsmath,amssymb,amsfonts}
\usepackage{algorithmic}
\usepackage{graphicx}
\usepackage{textcomp}
\usepackage{xcolor}
\usepackage{xurl}
\usepackage{hyperref}
\usepackage{fancybox}
\hypersetup{
    colorlinks=true,
    linkcolor=blue,
    filecolor=black,
    citecolor = blue,      
    urlcolor=blue,
}

\usepackage[numbers,sort&compress]{natbib}
\bibpunct[, ]{[}{]}{,}{n}{,}{,}

\theoremstyle{plain}

\theoremstyle{definition}

\theoremstyle{remark}

\begin{document}


\title{Using LLM for Real-Time Transcription and Summarization of Doctor-Patient Interactions into ePuskesmas in Indonesia: A Proof-of-Concept Study}

\author{
\name{Nur Ahmad Khatim\textsuperscript{a}, Ahmad Azmul Asmar Irfan\textsuperscript{b}\thanks{CONTACT ahmad.azmul@uinjkt.ac.id} and Mansur M. Arief\textsuperscript{c}\thanks{Code available at \url{https://github.com/mansurarief/LLM_ePuskesmas}}}
\affil{\textsuperscript{a}Institut Teknologi Sepuluh Nopember, Surabaya, Indonesia}
\affil{\textsuperscript{b}UIN Syarif Hidayatullah, Jakarta, Indonesia}
\affil{\textsuperscript{c}Stanford University Stanford, USA}
}

\maketitle

\begin{abstract}
One of the critical issues contributing to inefficiency in Puskesmas (Indonesian community health centers) is the time-consuming nature of documenting doctor-patient interactions. Doctors must conduct thorough consultations and manually transcribe detailed notes into ePuskesmas electronic health records (EHR), which creates substantial administrative burden to already overcapacitated physicians. This paper presents a proof-of-concept framework using large language models (LLMs) to automate real-time transcription and summarization of doctor-patient conversations in Bahasa Indonesia. Our system combines Whisper model for transcription with GPT-3.5 for medical summarization, implemented as a browser extension that automatically populates ePuskesmas forms. Through controlled roleplay experiments with medical validation, we demonstrate the technical feasibility of processing detailed 300+ seconds trimmed consultations in under 30 seconds while maintaining clinical accuracy. This work establishes the foundation for AI-assisted clinical documentation in resource-constrained healthcare environments. However, concerns have also been raised regarding privacy compliance and large-scale clinical evaluation addressing language and cultural biases for LLMs.
\end{abstract}


\vspace{-2em}
\section{Introduction}

In Indonesia, primary healthcare delivery through Community Health Centers (Puskesmas) faces mounting pressure from increasing patient volumes and administrative burdens. The Indonesian government has implemented ePuskesmas, a national electronic health record (EHR) system, to standardize patient data management across more than 10,000 Puskesmas facilities nationwide \cite{ePuskesmasBangkaBarat2024, demaio2014primary}. ePuskesmas is designed to help doctors and other healthcare providers manage patient data, ensuring easier monitoring of the patient health data for the Health Department at scale \cite{magdalena2023tinjauan}. However, the manual documentation requirements have created unexpected inefficiencies, with healthcare providers spending substantial time on data entry rather than direct patient care.

\begin{figure}
    \centering
    \includegraphics[width=\linewidth]{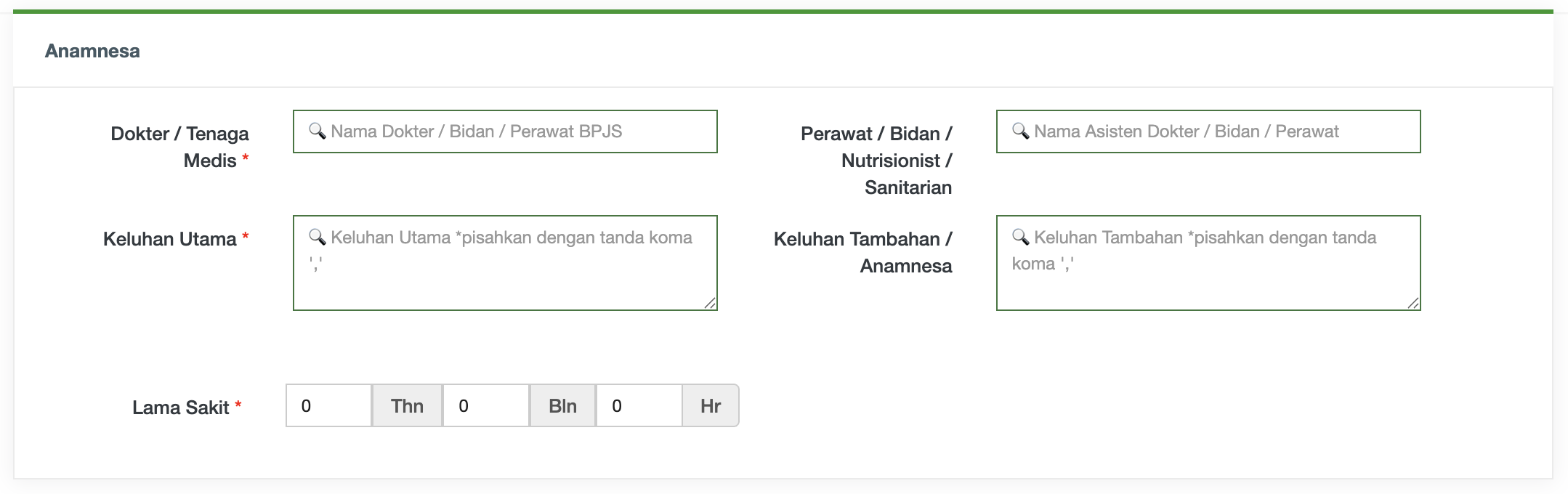}
    \includegraphics[width=\linewidth]{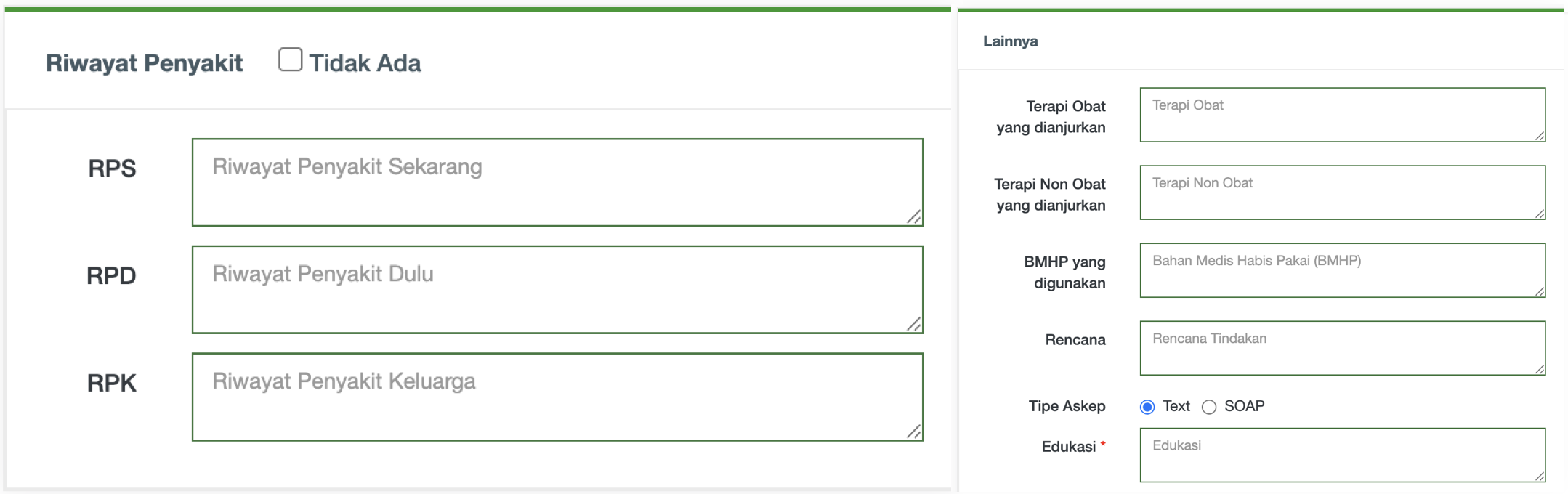}
    \includegraphics[width=\linewidth]{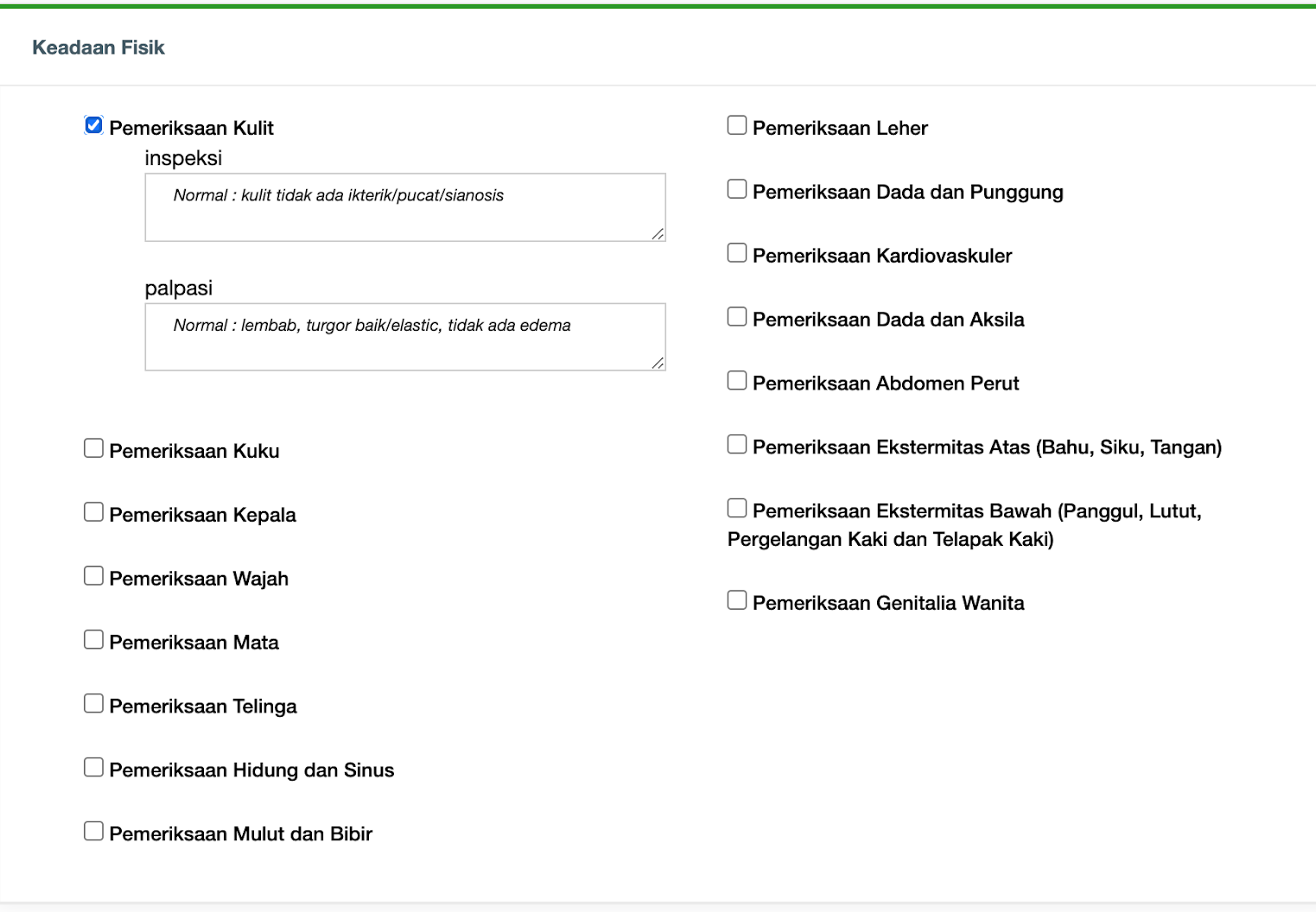}
    \caption{Current ePuskesmas interface requiring manual completions. Our framework automatically fills out all these fields from audio recording.}
    \label{fig:anamnesis_ui_1}
    \vspace{-1em}
\end{figure}

The documentation burden in Indonesian primary care is particularly acute due to dual recording requirements—providers must complete both digital ePuskesmas entries and maintain paper backup records \cite{oktora2018effectiveness}. 
Such paperwork and redundant systems often fall directly on providers, further increasing their administrative workload
\cite{braa2017health}.
Research indicates that manual EHR data entry contributes significantly to physician burnout globally \cite{sinsky2016allocation}, with studies showing that clinicians spend 1.67 times the consultation duration on documentation tasks \cite{yang2022large}. In resource-constrained environments like Indonesian Puskesmas, this administrative overhead directly reduces patient access to care by limiting provider availability. \autoref{fig:anamnesis_ui_1} shows the current UI of the anamnesis page of ePuskesmas, which is currently manually filled out by the doctor.

Recent advances in large language models (LLMs) present promising opportunities for clinical documentation automation. OpenAI's Whisper model has demonstrated robust multilingual transcription capabilities, including support for Bahasa Indonesia \cite{radford2023robust}, while GPT-3.5 has shown effectiveness in medical text summarization tasks \cite{van2024adapted}. However, most existing research focuses on high-resource healthcare settings with reliable internet connectivity and established data governance frameworks, which are not the conditions in Indonesian primary care facilities.

This paper presents a proof-of-concept framework that addresses the specific constraints of Indonesian healthcare through several key design decisions: (1) browser-based implementation for easy deployment across diverse Puskesmas IT environments, (2) structured prompting optimized for Bahasa Indonesia medical terminology, and (3) direct integration with existing ePuskesmas workflows. Our system, illustrated in \autoref{fig:proposed_framework}, combines real-time audio capture, automated transcription and summarization, and systematic form population. 

Our work makes three primary contributions within the scope of proof-of-concept research. \textbf{First}, we develop and demonstrate an end-to-end technical framework for automated transcription and summarization of doctor-patient interactions in Bahasa Indonesia, specifically designed for integration with Indonesia's national ePuskesmas EHR system. Our browser-based approach addresses deployment constraints common in resource-limited healthcare environments. \textbf{Second}, we provide initial validation of technical feasibility through controlled roleplay experiments with medical expert review, demonstrating successful processing of typical primary care consultations in under 30 seconds while maintaining clinically appropriate information extraction. \textbf{Third}, we conduct comprehensive analysis of deployment limitations, infrastructure requirements, and privacy considerations specific to LMIC healthcare contexts, providing a roadmap for future research toward clinical implementation. However, we emphasize that the current work represents proof-of-concept study aiming to assess the feasibility of the proposed framework and identify the challenges in the context of resource-constrained LLMs for healthcare applications, which requires substantial additional research before large-scale clinical implementation.

The rest of this study is organized as follows. In Section \ref{literature_review}, we review related research. In Section \ref{methodology_and_system_framework}, we detail our technical framework design. Then, Section \ref{results} presents our controlled experimental methodology and validation results from two roleplay scenarios with medical expert review. Section \ref{discussion} highlights some of our preliminary findings and discusses the limitations and concerns of the current work. Finally, Section \ref{conclusion} summarizes our conclusions. The code for the prototype is available at \url{https://github.com/mansurarief/LLM_ePuskesmas}.

\section{Literature Review}
\label{literature_review}

We briefly review existing work on LLM applications in clinical documentation, multilingual healthcare AI, and infrastructure challenges specific to LMIC healthcare deployment

\subsection{Clinical Documentation Automation with LLMs}

Healthcare documentation automation has emerged as a critical application area for large language models, with several studies demonstrating promising results in controlled settings. Van Veen et al. \cite{van2024adapted} conducted one of the most comprehensive evaluations, showing that fine-tuned LLMs could match or exceed physician performance in clinical summarization tasks across radiology reports, progress notes, and patient dialogues. Their study involved 26 physician evaluators and found LLM summaries rated as equivalent (45\%) or superior (36\%) to expert-generated summaries. The GatorTron model represents another significant advancement, achieving F1 scores of 0.8996 for medical concept extraction and 0.9627 for relation extraction on clinical NLP benchmarks \cite{yang2022large}. Notably, GatorTron's automated summaries required 1.67 times less completion time than human scribes while maintaining superior accuracy ratings. However, these systems were evaluated primarily on English-language datasets with high-resource healthcare settings.

For EHR automation specifically, Li et al. \cite{li2024scoping} identified seven primary LLM application areas: named entity recognition, information extraction, text summarization, clinical decision support, medical coding, patient communication, and research data extraction. Their scoping review found strongest evidence for text summarization applications, though most studies lacked rigorous clinical validation protocols. A critical counterpoint comes from Wornow et al. \cite{wornow2023shaky}, who examined 84 foundation models trained on EMR data and found that most relied on narrow datasets or overly broad biomedical corpora. Their analysis suggests that many current approaches may not generalize effectively to diverse clinical environments, particularly in resource-constrained settings with different documentation practices and terminology.

\subsection{Multilingual and Low-Resource Healthcare AI}

The application of LLMs in non-English healthcare contexts remains limited but growing. Wang et al. \cite{wang2024can} demonstrated Whisper's effectiveness for Chinese dialect transcription using speech-based in-context learning, achieving significant error reductions through language-level adaptation techniques. Similarly, Jelassi et al. \cite{jelassi2024revolutionizing} showed successful deployment of Whisper for French radiological report transcription, though their study was limited to a single medical specialty. For Bahasa Indonesia specifically, research on medical LLM applications is extremely limited. Most existing work focuses on general-purpose language tasks rather than specialized medical applications. This gap represents a significant limitation for healthcare AI deployment in Indonesia, where medical terminology, consultation patterns, and documentation requirements may differ substantially from Western healthcare systems.

\subsection{Infrastructure and Deployment Challenges in LMICs}

Healthcare AI deployment in Low- and Middle-Income Countries (LMICs) faces distinct challenges that are often underestimated in research literature. Ciecierski-Holmes et al. \cite{ciecierski2022artificial} conducted a systematic review of AI implementation in LMIC healthcare settings, identifying infrastructure limitations as the primary barrier to successful deployment. Power and internet connectivity represent fundamental constraints. Healthcare facilities in LMICs experience system outages 2-3 times more frequently than high-income countries, with many rural facilities lacking reliable broadband access \cite{zuhair2024exploring}. These connectivity issues are particularly problematic for cloud-based LLM systems that require stable internet for API access.

Equipment sustainability poses another critical challenge. Studies indicate that 70-90\% of donated medical equipment in LMICs fails within five years due to inadequate maintenance, lack of spare parts, or insufficient technical training \cite{alami2020artificial}. This pattern suggests that healthcare AI systems must be designed for minimal maintenance requirements and local technical capacity. Data governance and privacy compliance represent additional complexity layers. Many LMICs lack comprehensive healthcare data protection regulations equivalent to HIPAA or GDPR, creating uncertainty about appropriate data handling practices for AI systems processing sensitive patient data.

\begin{figure*}
    \centering
    \includegraphics[width=\linewidth]{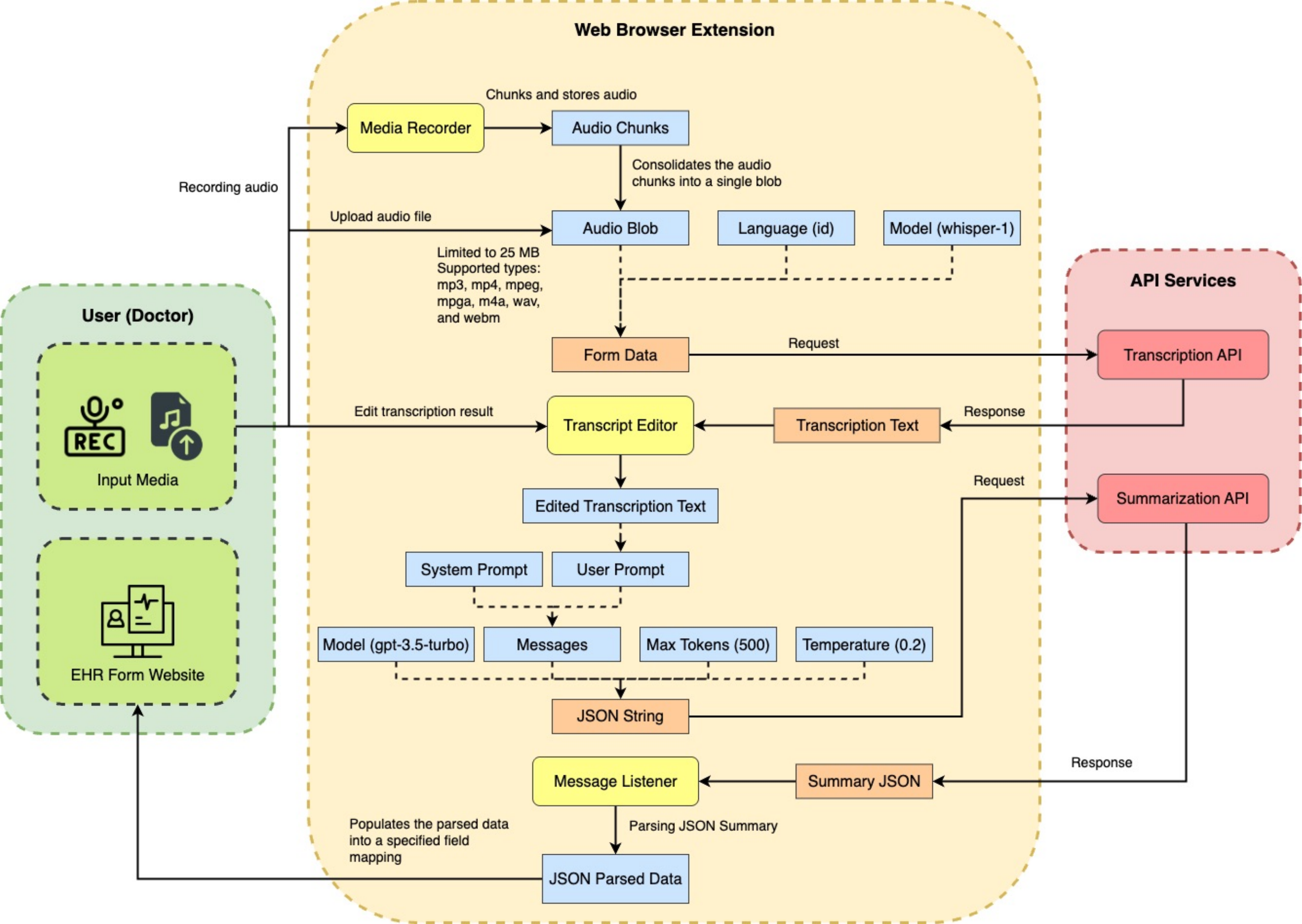}
    \caption{Proposed LLM-based framework for automated clinical documentation in ePuskesmas. The system integrates browser-based audio capture, cloud-based transcription/summarization, and automated form population while highlighting critical dependency on stable internet connectivity.}
    \label{fig:proposed_framework}
\end{figure*}

\section{Proposed Framework}
\label{methodology_and_system_framework}

In this study, we address several critical gaps: (1) limited research on clinical documentation automation in Bahasa Indonesia, (2) insufficient attention to deployment constraints in resource-limited healthcare settings, and (3) lack of integration between multilingual transcription capabilities and existing EHR workflows in LMICs. These issues are crucial for the next phase of larger clinical validation in resource-constrained and diverse settings such as Indonesia's Puskesmas. 

Our framework addresses the specific constraints of Indonesian primary healthcare through a browser-based approach that integrates with existing ePuskesmas workflows. The system architecture, shown in \autoref{fig:proposed_framework}, prioritizes deployment simplicity and compatibility with diverse IT environments commonly found in Puskesmas facilities.

\subsection{Design Principles and Constraints}

The framework design is guided by several key constraints identified through preliminary consultation with Puskesmas staff: (1) minimal additional hardware requirements, (2) compatibility with existing browser-based ePuskesmas access, (3) minimal disruption to established consultation workflows, and (4) accommodation of variable internet connectivity quality. We chose a browser extension approach over standalone applications to leverage existing IT infrastructure and reduce deployment complexity. However, this design decision creates inherent limitations, particularly regarding offline functionality and local data processing capabilities, which we address in our discussion of limitations.

\subsection{Conversation Recording}

The system initiates with browser-based audio capture using JavaScript's MediaRecorder API \cite{MDNWebDocsMediaRecorder}. The extension provides simple controls (start/stop recording, upload existing files) to minimize workflow disruption during patient consultations. Audio data is captured in real-time and stored locally in chunks until processing initiation. Our implementation showcases this functionality using Google Chrome for simplicity, but can be easily adapted to other browsers by using their corresponding function calls. Audio is currently captured at 44.1 kHz sampling rate and compressed using the browser's default audio codec (typically WebM or MP3). The system automatically handles audio format conversion for Whisper API compatibility, resampling to the required 16 kHz format during preprocessing. The current implementation lacks noise reduction, speaker diarization, or audio quality assessment—features that has been identified as essential in busy healthcare facilities such as Puskesmas where background noise common.

\subsection{Transcription with Whisper}

We utilize OpenAI's Whisper model through their API service for speech-to-text conversion. Whisper was selected for its demonstrated multilingual capabilities and specific support for Bahasa Indonesia, though we acknowledge that performance may vary significantly across Indonesian dialects and regional linguistic variations. Moreover, Whisper's performance on medical terminology in Bahasa Indonesia has not been systematically evaluated. Our preliminary testing suggests adequate performance for common medical terms, but accuracy may decline for specialized terminology or when handling multiple speakers simultaneously. For our implementation, we leveraged the OpenAI API to access the Whisper model. While this approach allowed us to integrate the model seamlessly into our system, benefiting from its advanced capabilities without the need for extensive local infrastructure, it requires a stable internet connection to interface with OpenAI's servers and obtain results, ensuring that we stay updated with the latest advancements in speech processing technology. It processes raw audio signals re-sampled to 16,000 Hz and outputs text transcriptions directly.

\subsection{Summarization with GPT-3.5}

The summarization component employs GPT-3.5-turbo with carefully structured prompts designed for Indonesian medical context. Our prompt engineering process involved iterative refinement based on consultation with Indonesian physicians to ensure appropriate medical terminology and cultural sensitivity. The system prompt instructs the model to act as an expert medical transcriber familiar with Indonesian healthcare terminology and ePuskesmas documentation requirements. We specify eight structured output categories: Chief Complaint, Additional Complaints, History of Present Illness, Past Medical History, Family History, Recommended Medication Therapy, Non-Medication Therapy, and Education. The model generates structured JSON output with predefined keys corresponding to ePuskesmas form fields. We set \texttt{max\_tokens} to 500 and \texttt{temperature} to 0.2 to encourage focused, consistent outputs while maintaining sufficient detail for clinical utility. The current summarization approach does not include validation mechanisms for medical accuracy or clinical appropriateness. Generated summaries may contain factual errors or inappropriate clinical interpretations that could impact patient care if used without human oversight. We rely on medical expert validation to ensure the accuracy and clinical relevance of the generated summaries.

\subsection{ePuskesmas Integration}

\begin{figure}
    \centering
    \includegraphics[width=\linewidth]{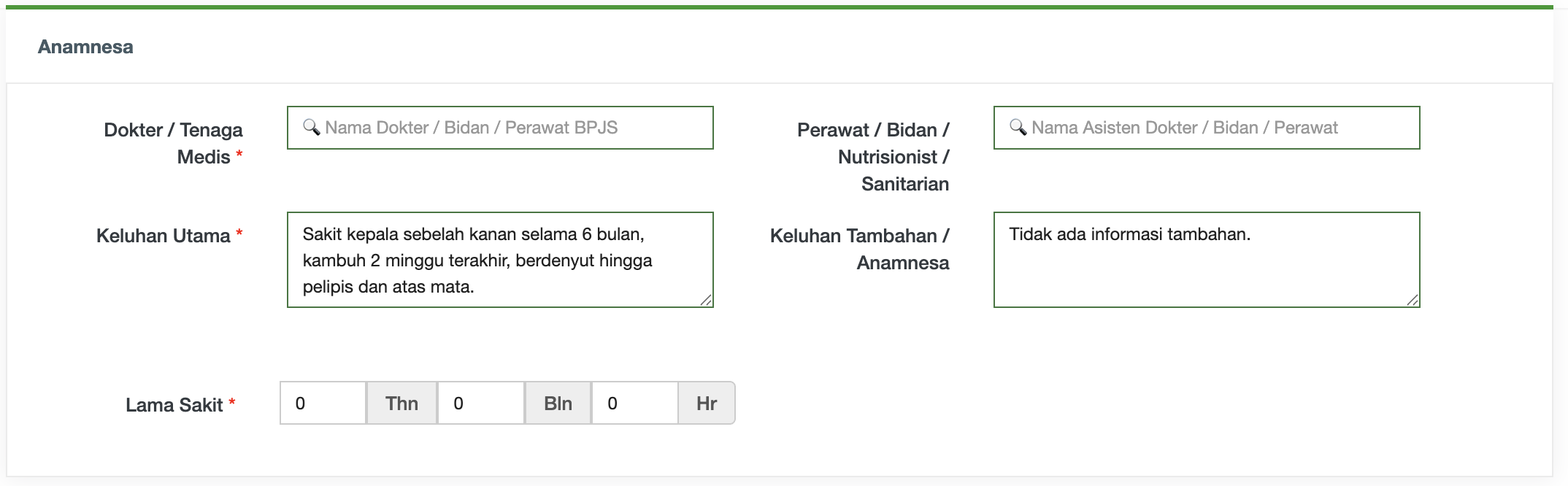}
    \caption{Automated population of primary complaint fields in ePuskesmas interface, demonstrating successful system integration.}
    \label{fig:form_parsed_exp2}
\end{figure}

\begin{figure}
    \centering
    \includegraphics[width=\linewidth]{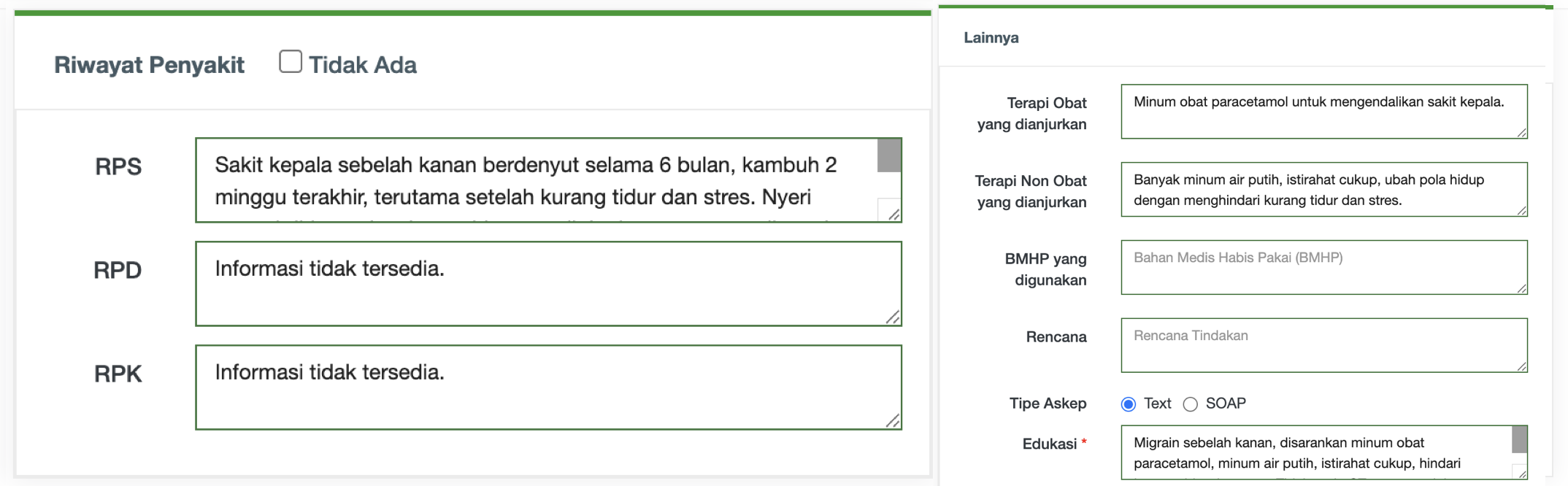}
    \caption{Complete automated form population including history, therapy recommendations, and education components.}
    \label{fig:form_parsed_exp2_2}
\end{figure}

\begin{figure}
    \centering
    \includegraphics[width=\linewidth]{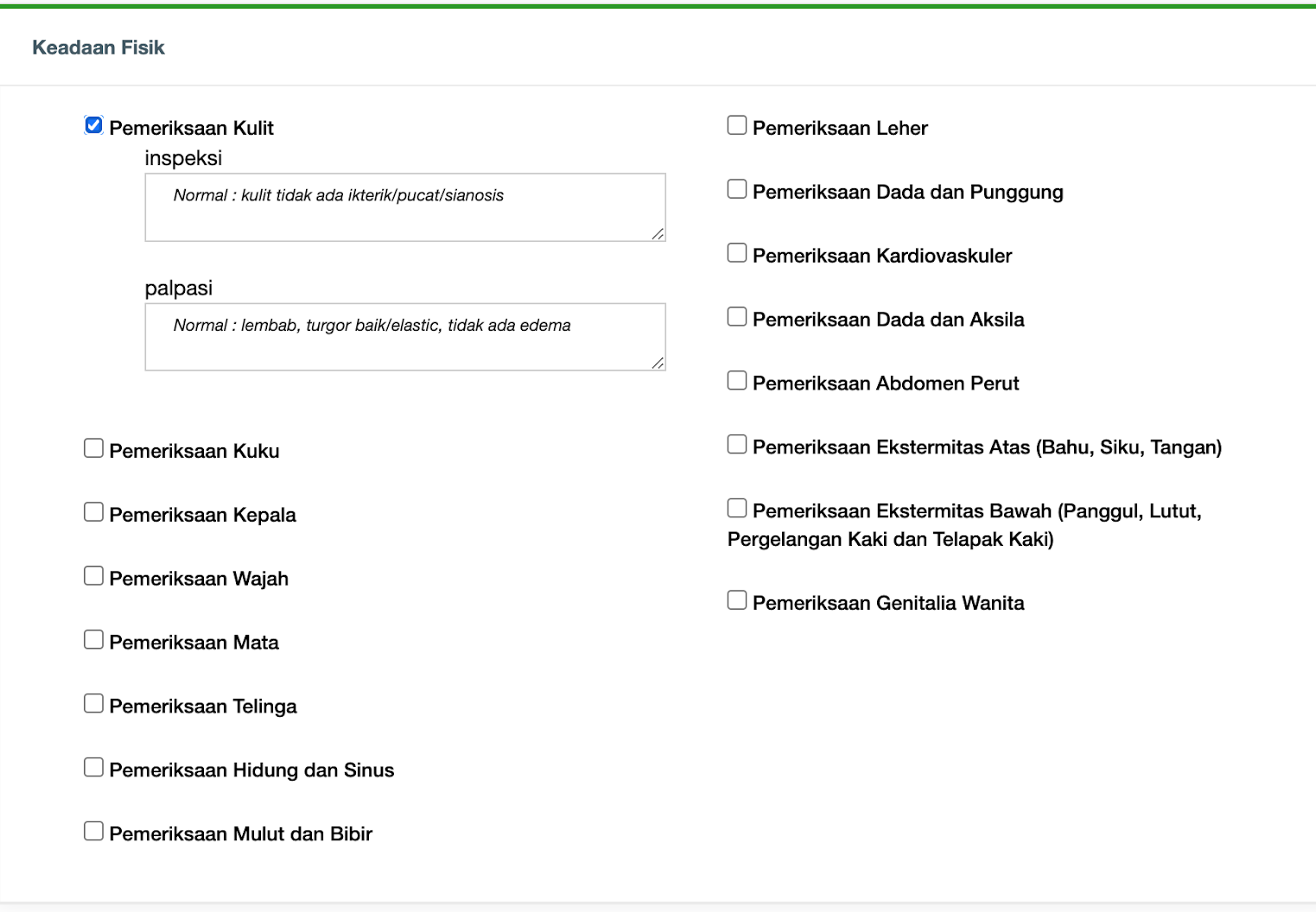}
    \caption{Physical examination section of ePuskesmas currently requiring manual completion, highlighting a critical limitation of the conversation-based approach.}
    \label{fig:anamnesis_ui_finetuning}
\end{figure}

The final component automatically populates ePuskesmas forms using JavaScript DOM manipulation. The system maps JSON output keys to specific form field identifiers, enabling direct data insertion without manual transcription. Field mapping is achieved through predefined identifier matching between JSON keys (\texttt{chief\_complaint}, \texttt{additional\_complaint}, etc.) and ePuskesmas form input IDs. The system includes error handling for missing fields or formatting inconsistencies. Examples of mapped data are shown in \autoref{fig:form_parsed_exp2}, \autoref{fig:form_parsed_exp2_2}, and \autoref{fig:anamnesis_ui_finetuning}. Current implementation is specifically tailored to the ePuskesmas interface version tested, and hence it is not generalizable to other ePuskesmas interfaces. Form updates or interface changes would require corresponding system modifications for the current setup.

\subsection{Prototype}
\begin{figure}
    \centering
    \includegraphics[width=0.5\linewidth]{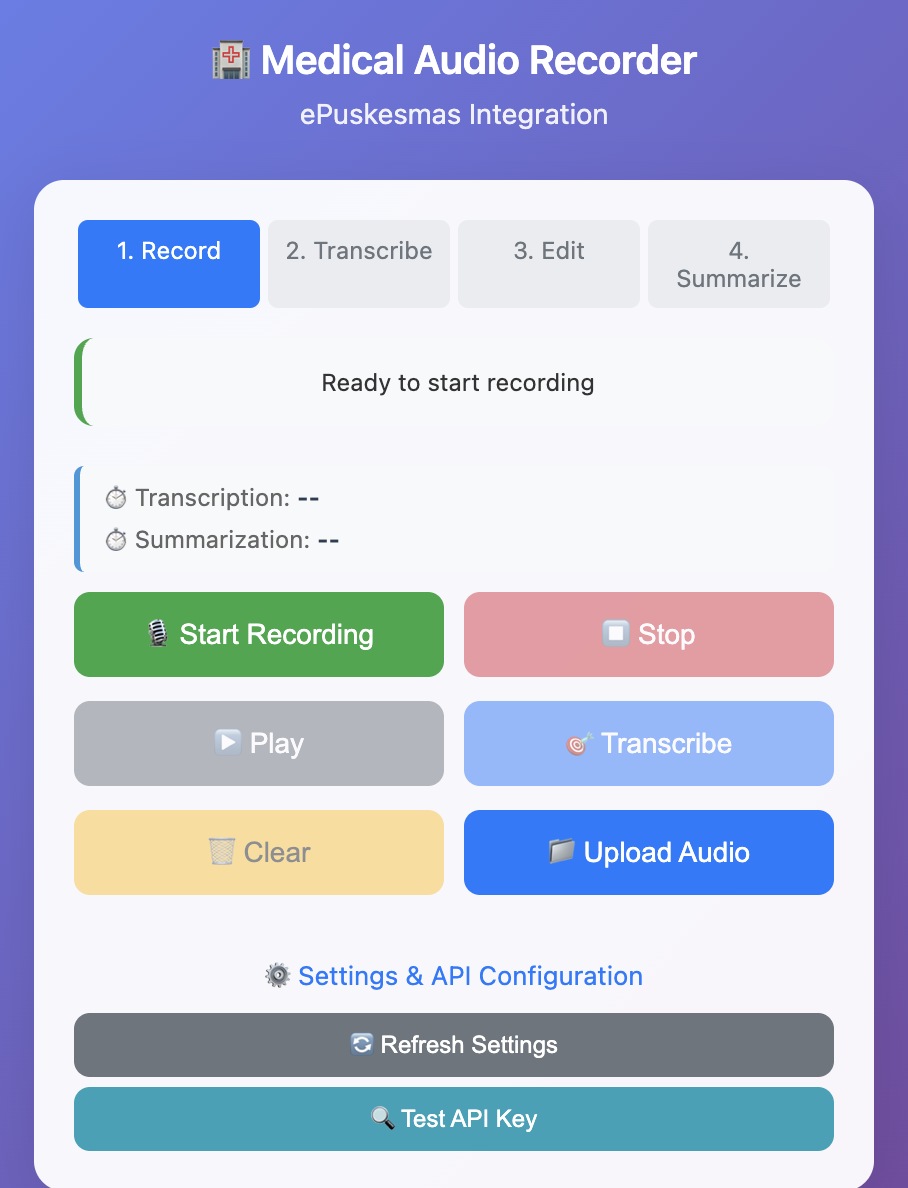}
    \caption{Browser extension user interface during processing, designed for minimal workflow disruption during clinical consultations.}
    \label{fig:extension_ui}
\end{figure}

At the current stage, we have developed a functional prototype in the form of a web browser extension. This extension has been successfully tested on a real ePuskesmas form, demonstrating its end-to-end functionality. The extension is capable of recording audio, calling the corresponding APIs to perform both transcription and summarization using the Whisper model and GPT-3, and subsequently populating the form's input fields with the generated summaries. The added benefit of this prototype is that it can be deployed in any ePuskesmas application that is mainly accessed through a web browser, not just the one in Bangka Barat. The prototype while processing transcription and summarization is shown in \autoref{fig:extension_ui}. As will be discussed later, based on our testing environment, the complete processing pipeline (transcription + summarization + form population) requires 20-30 seconds for 5-6 minute consultations. Network latency represents the primary bottleneck, with API calls accounting for 80-90\% of total processing time.

\section{Experiment \& Results} \label{results}

Our experiments employed controlled roleplay scenarios with medical validation to assess system functionality and clinical appropriateness. Each experiment was reviewed by a practicing physician from Puskesmas Tempilang who assessed: (1) transcription accuracy for medical terminology, (2) clinical appropriateness of extracted information, (3) completeness of documentation relative to consultation content, and (4) suitability for ePuskesmas workflow integration. Our evaluation is limited to two scenarios with single-validator assessment. This approach lacks the statistical power and inter-rater reliability measures required for robust clinical validation. Additionally, roleplay scenarios may not capture the complexity, interruptions, and acoustic challenges of real clinical environments, which is highlighted in our discussions.

While this approach enables systematic testing under controlled conditions, by no means it represents a significant limitation compared to real-world clinical validation. Our goal is to obtain early insights from users for our proof-of-concept. Substantial clinical validation should follow prior to large-scale deployments, which we defer in our future work.

\subsection{Scenario 1: Gastrointestinal Complaint (Dyspepsia)}

The first experiment simulated a common primary care presentation: dyspepsia (locally termed ``maag'')
\cite{syam2017national}. This condition represents 30\% of services provided by general physicians in Indonesian studies, making it representative of routine clinical encounters
\cite{darnindro2020prevalence}. The audio quality was good, with minimal background noise. The doctor was also able to speak clearly without any interruptions. The patient was able to speak Bahasa Indonesia with good understanding of standard medical terminology.

The effective conversation lasted for 332 seconds. The full conversation was transcribed and the corresponding text output shown in \autoref{fig:transcription_text_exp1} and the corresponding JSON output shown in \autoref{fig:summarization_json_exp1}. Total processing time was 28 seconds (transcription: 23 seconds, summarization: 4 seconds, form population: 1 second). The system successfully extracted key clinical elements including primary complaint, symptom duration, lifestyle triggers, and treatment recommendations. The reviewing physician confirmed accuracy of extracted information and appropriateness of clinical categorization. Notable strengths included correct identification of stress-related triggers and appropriate medication recommendations. The extracted data was then successfully parsed and automatically populated into the ePuskesmas form. However, some nuanced discussion about symptom severity was incompletely captured. 

\begin{figure*}[htbp!]
\boxed{
\begin{minipage}{0.98\linewidth}
\scriptsize
\textit{}Assalamualaikum Bu, good morning, please sit down. I am Dr. Azmul who works at this puskesmas. This morning with Mrs... what is your name? Good morning Mrs. Novi Mrs. Novi, how old are you Bu? 40 years old, so I have been called Mrs. Mbak, what's wrong? What does Mrs. Novi have complaints about this morning? What's wrong Mbak? Yes doctor, my stomach has been hurting for about 3 days now, the pain in my stomach, when will it heal doctor? What does the pain feel like Mrs. Novi? Can you describe what kind of pain it is? Sometimes it feels hot, sometimes it feels like there are stomach acid symptoms or something like that, doctor. The pain, what triggers it? Is there anything that makes it worse? Sometimes when I'm hungry and haven't eaten, doctor. Oh, when you're hungry. Usually when does it hurt the most? During the day when busy with work or what? Sometimes during the day, but it also hurts at night, doctor. During the night too. Has Mrs. Novi taken any medication? What has Mrs. Novi taken? I have taken stomach medicine, doctor. What kind of stomach medicine have you taken? Sometimes Promag, sometimes other stomach medicines. How about the results after taking the medicine? The pain reduced a bit, but it came back again doctor. The pain came back. How often does this stomach pain occur? Once a month? Once a week? It depends doctor, sometimes once a week, sometimes twice a week, depends on my activities. What about eating patterns? What does Mrs. Novi usually eat? I often eat late, doctor. Often eat late. What time do you usually eat? Sometimes at 10, sometimes at 11, sometimes at 2 in the afternoon when I'm really hungry, then I eat. What foods do you often eat? Spicy foods, fried foods. Do you often eat fried foods too? Yes, I like spicy and fried foods doctor. How about drinks? What do you usually drink? Water, sometimes coffee. Coffee, how often do you drink coffee? Every day, doctor. Every day, how many times? About 2-3 times a day. That's quite frequent. Do you add sugar to your coffee? Yes, I add sugar. How many spoons of sugar? About 2 spoons doctor. Two spoons. Do you often feel stressed Mrs. Novi? Do you feel stressed or worried about something? Sometimes I feel stressed because of work, doctor. Work stress. What kind of work do you do Mrs. Novi? I work in an office. Office work. How many hours do you work per day? About 8 hours, doctor. 8 hours. Do you often work overtime? Sometimes, doctor. Do you often skip meals because of work? Yes, often doctor, because of deadlines I often don't have time to eat. Because of deadlines. Have you had this stomach pain before Mrs. Novi? Have you experienced this before? Yes, I have experienced this several times, doctor. Several times. When was the last time before this? Last month, doctor. Last month, how long did it last? About 1 week, doctor. 1 week. Did you go to the doctor at that time? Yes, I went to the doctor. What did the doctor say? The doctor said it was stomach acid, doctor. Stomach acid. What medicine did the doctor give? The doctor gave stomach medicine, antacids, doctor. How long did you take the medicine? About 1 week, doctor. 1 week, did you finish all the medicine? Yes, I finished it all, doctor. Did the complaints disappear after that? Yes, they disappeared, doctor, but came back after about 2 weeks. They disappeared then came back. Mrs. Novi, do you smoke? No, I don't smoke, doctor. Don't smoke. How about alcohol? I don't drink alcohol either, doctor. Don't drink alcohol. Do you often eat late at night? Sometimes, doctor. What do you usually eat at night? Sometimes instant noodles, sometimes fried foods, doctor. Do you have family members who have stomach problems like this? My mother has stomach problems too, doctor. Your mother also has stomach problems. What stomach problems does your mother have? Stomach acid, doctor. Stomach acid. Is your mother still being treated for it? Yes, still being treated, doctor. Still being treated. Mrs. Novi, based on our conversation, your stomach pain seems to be related to your eating patterns and stress, do you agree? Yes, I agree doctor. So the main advice I want to give is about eating patterns. Try to eat regularly, 3 times a day at the same time. Don't skip meals even when busy with work. For medicines, I will give you stomach medicine to reduce stomach acid, also medicine to protect your stomach lining. Also try to reduce spicy and fried foods, reduce coffee, maybe just 1 cup a day. Try to manage stress, if possible do light exercise, get enough rest. If the pain doesn't improve in 1 week, please come back for another check-up. Do you understand Mrs. Novi? Yes, I understand doctor, thank you very much.
\end{minipage}
}
\caption{Complete English translation transcription output for gastrointestinal complaint scenario, demonstrating the framework's capability with Indonesian medical terminology. Original version is provided in Appendix.}
\label{fig:transcription_text_exp1}
\end{figure*}

\begin{figure*}[htbp!]
\boxed{
\begin{minipage}{0.98\linewidth}
\scriptsize
\{\\
\hspace*{1em}``chief\_complaint'': ``Stomach pain for the past 3 days, burning sensation and stomach acid symptoms'',\\
\hspace*{1em}``additional\_complaint'': ``Bloating, burning sensation, abdominal pain when hungry'',\\
\hspace*{1em}``history\_of\_present\_illness'': ``Stomach pain for the past 3 days, burning sensation when stomach is empty. Pain occurs both day and night. Has taken stomach medication, \\
\hspace*{4em}pain temporarily reduced but returned'',\\
\hspace*{1em}``past\_medical\_history'': ``No information available'',\\
\hspace*{1em}``family\_history'': ``No information available'',\\
\hspace*{1em}``recommended\_medication\_therapy'': ``Stomach acid medication: Omeprazole for stomach acid reduction, Antacid for stomach protection'',\\
\hspace*{1em}``recommended\_non\_medication\_therapy'': ``Regular eating pattern, reduce spicy and fried foods, reduce coffee consumption, stress management'',\\
\hspace*{1em}``education'': ``Avoid stress, reduce spicy foods, reduce coffee consumption, regular eating pattern, sufficient rest. Return if no improvement after taking medication for 1 week''\\
\}
\end{minipage}
}
\caption{Structured English translation JSON summary output for Scenario 1, showing systematic extraction of clinical information into ePuskesmas-compatible format. Original version is provided in Appendix.}
\label{fig:summarization_json_exp1}
\end{figure*}

\subsection{Scenario 2: Neurological Complaint (Migraine)}

The second experiment evaluated system performance on a more complex neurological presentation: recurring migraine headaches. This scenario was selected to test the system's ability to handle symptom complexes and temporal patterns common in chronic conditions. The clinical complexity was considered moderate, with multiple triggers. The language used was also more complex, with the patient using more nuanced conversation patterns.

The trimmed conversation lasted for 384 seconds, with the audio accurately converted into text, with the transcription result shown in \autoref{fig:transcription_text_exp2}, with the corresponding JSON output shown in \autoref{fig:summarization_json_exp2}. Processing completed in 28 seconds (transcription: 25 seconds, summarization: 2 seconds, form population: 1 seconds). The system demonstrated improved summarization efficiency for the longer consultation, possibly due to clearer temporal structure in the migraine presentation. Validation confirmed appropriate extraction of headache characteristics, trigger identification, and treatment plan documentation. The system successfully distinguished between patient-reported symptoms and physician-recommended interventions, which yields a complete and well-structured output. The extracted data was then successfully parsed and automatically populated into the ePuskesmas form, as shown in \autoref{fig:form_parsed_exp2_2}.

\begin{figure*}[htbp!]
\boxed{
\begin{minipage}{0.98\linewidth}
\scriptsize
\textit{}Assalamualaikum, good morning, please sit down Mrs... I am Dr. Asmul who works at this Puskesmas. This morning with Mrs... what is your name? Good morning, I am Mrs. Desi Oh Mrs. Desi Mrs. Desi, how can I help you Mrs? Yes doctor, I want to complain that I have been having headaches for the past 6 months, every 2 weeks the headache comes back, along with nausea. The headaches have been going on for 6 months, every 2 weeks the headache comes back. What does the headache feel like Mrs? It throbs on the left side of my head. Throbbing on the left side. How severe is the pain on a scale of 1 to 10? Around 7-8, doctor. 7-8, that's quite severe. How long does each headache episode usually last? Usually around 4-8 hours, doctor. 4-8 hours. When the headache comes, are there any warning signs before the headache starts? Sometimes I see flashing lights, doctor. You see flashing lights. What triggers your headaches? I notice when I'm stressed, lack of sleep, or when the weather is too bright. Stress, lack of sleep, bright weather. Have you taken any medication for these headaches? I have taken paracetamol, doctor. Paracetamol. How effective is the paracetamol? Sometimes it helps, but often the pain doesn't completely go away, doctor. Doesn't completely go away. Besides paracetamol, have you tried other treatments? I haven't tried other treatments, doctor. When you have a headache, do you prefer to be in a dark or bright place? I prefer dark places, doctor. Dark places. How about sound? Do you prefer quiet or can you tolerate noise? I prefer quiet places, doctor, because noise makes the headache worse. Noise makes it worse. You mentioned nausea, do you also vomit? Sometimes I do vomit, doctor. Sometimes you vomit. Do you have a family history of headaches like this? My mother also often has headaches, doctor. Your mother also has headaches. Does your mother take any specific medication for her headaches? My mother takes medication from the doctor, but I don't know what kind, doctor. You don't know what medication. How often do you experience stress Mrs. Desi? Quite often, doctor, because of work and taking care of children. Work stress and taking care of children. What is your job? I work as a teacher, doctor. Teacher. How many hours of sleep do you usually get? Around 5-6 hours, doctor. 5-6 hours, that's quite little. Do you often stay up late? Yes, doctor, I often stay up late preparing lessons or checking student work. Stay up late for lesson preparation. Do you drink coffee or tea regularly? I drink coffee almost every day, doctor. Coffee every day. How many cups per day? Around 2-3 cups, doctor. 2-3 cups. Do you think there's a connection between your coffee consumption and your headaches? I'm not sure, doctor, but sometimes when I don't drink coffee I also get headaches. When you don't drink coffee you also get headaches. Have you tried to identify other triggers for your headaches? I notice that when I'm very tired or when I skip meals, the headaches often come, doctor. When very tired or skipping meals. Do you exercise regularly? I rarely exercise, doctor. Rarely exercise. Mrs. Desi, based on our conversation, it seems like you are experiencing migraine headaches. The characteristics you describe - one-sided throbbing pain, nausea, sensitivity to light and sound, and seeing flashing lights - are typical symptoms of migraine. The triggers you mentioned like stress, lack of sleep, and bright light are also common migraine triggers. For treatment, I will give you pain medication specifically for migraines, not just paracetamol. I will also give you anti-nausea medication. It's important to get adequate rest, try to sleep 7-8 hours per night, manage stress well, avoid headache triggers you've identified. Try to eat regularly, don't skip meals. If the headaches don't improve or get worse, please come back for follow-up. Do you understand Mrs. Desi? Yes, I understand doctor, thank you very much.
\end{minipage}
}
\caption{Complete English translation transcription for migraine consultation scenario, illustrating system performance with complex symptom presentations. Original version is provided in Appendix.}
\label{fig:transcription_text_exp2}
\end{figure*}

\begin{figure*}[htbp!]
\boxed{
\begin{minipage}{0.98\linewidth}
\scriptsize
    \{\\
    \hspace*{1em}``chief\_complaint'': ``Headache for the past 6 months, every 2 weeks for the past 6 months, every 2 weeks recurring, along with nausea and sometimes vomiting'',\\
    \hspace*{1em}``additional\_complaint'': ``No additional information'',\\
    \hspace*{1em}``history\_of\_present\_illness'': ``Headache for the past 6 months recurring every 6 months, every 2 weeks recurring, especially after stress, lack of sleep and bright light, usually \\
    \hspace*{4em}lasting about 4-8 hours'',\\
    \hspace*{1em}``past\_medical\_history'': ``No information available'',\\
    \hspace*{1em}``family\_history'': ``No information available'',\\
    \hspace*{1em}``recommended\_medication\_therapy'': ``Migraine-specific pain medication to control headache'',\\
    \hspace*{1em}``recommended\_non\_medication\_therapy'': ``Adequate rest, minimum 7-8 hours sleep, stress management, adequate sleep, avoid bright light, regular eating, avoid skipping \\
    \hspace*{4em}meals'',\\
    \hspace*{1em}``education'': ``Migraine education, stress reduction, adequate sleep, avoid bright light, regular meals. Don't need CT scan for now''\\
    \}
\end{minipage}
}
\caption{Structured English translation summary output for migraine scenario, demonstrating comprehensive clinical information extraction and categorization. Original version is provided in Appendix.}
\label{fig:summarization_json_exp2}
\end{figure*}

\section{Discussion} \label{discussion}

\subsection{Contextualizing the Results}

In the larger scope of the research, in hindsight, our results align with broader trends in healthcare AI research while highlighting unique challenges for LMIC deployment. Compared to studies like Van Veen et al. \cite{van2024adapted}, our work extends multilingual capabilities to Indonesian healthcare contexts but with significantly more limited evaluation scope. The infrastructure challenges we identify align with findings from Ciecierski-Holmes et al. \cite{ciecierski2022artificial} regarding AI deployment in LMICs. However, our work provides specific quantitative data on internet requirements and cost implications that were not previously available for clinical documentation automation in Indonesian settings. 

\subsection{Technical Feasibility and Performance Analysis}

Our proof-of-concept demonstrates the technical feasibility of automated clinical documentation for Indonesian primary care, with several notable achievements. The system successfully processed consultations lasting 5-6 minutes in under 30 seconds, representing a 10-fold improvement in documentation efficiency compared to manual entry. Both experimental scenarios achieved clinically appropriate information extraction with maintenance of medical terminology accuracy and logical clinical categorization. The integration with existing ePuskesmas infrastructure represents a significant practical advancement. Unlike previous research focusing on standalone applications or specialized hardware, our browser-based approach leverages existing IT capabilities while minimizing deployment complexity. This design decision enables potential implementation across Indonesia's diverse Puskesmas facilities without requiring substantial infrastructure investment. However, these promising results must be interpreted within the context of our limited experimental scope. Controlled roleplay scenarios cannot capture the acoustic complexity, interruption patterns, and terminology variation characteristic of real clinical environments. The acoustic environment in busy Puskesmas facilities includes significant background noise, multiple conversations, and equipment sounds that may substantially degrade transcription accuracy. The current system excels at capturing the patient's history but does not integrate findings from the physical examination, a crucial component of a comprehensive medical record as shown in \autoref{fig:anamnesis_ui_finetuning}.

\subsection{Clinical Documentation Quality}

The clinical validation revealed both strengths and limitations in our automated documentation approach. Positive aspects include accurate extraction of primary complaints, appropriate categorization of symptoms and treatments, and maintenance of clinical logic throughout the summarization process. The system demonstrated particular effectiveness in capturing education components and medication recommendations, which are the elements often inadequately documented in manual records due to time constraints. 

However, several clinical limitations became apparent during validation. Nuanced aspects of symptom severity and temporal patterns were occasionally oversimplified or incompletely captured. The system's inability to process physical examination findings represents a critical gap, as these components are essential for comprehensive clinical documentation and diagnostic accuracy. Furthermore, our current approach lacks validation mechanisms for clinical appropriateness or medical accuracy. This aligns with broader technical challenges observed in LLMs, including hallucination risks, insufficient clinical benchmarks, and the paucity of prospective validation trials. As highlighted in prior frameworks, linguistically coherent outputs may still be clinically inappropriate, underscoring the need for rigorous validation protocols, standardized reporting guidelines, and continuous human oversight to mitigate potential patient safety risks
\cite{jung2025large}. 
The system may generate summaries that are linguistically coherent but clinically inappropriate, potentially introducing documentation errors that could impact patient care. This limitation highlights the critical need for human oversight and validation protocols before any clinical deployment. Similar findings were noted in radiology, where GPT-4 achieved higher accuracy with in-context examples, emphasizing the need for contextual adaptation and validation, as technical fluency alone cannot ensure clinical appropriateness
\cite{kim2025context}.

\subsection{Deployment Constraints and Infrastructure Requirements}

Our analysis reveals significant deployment constraints that limit immediate real-world applicability. Internet connectivity requirements represent the most substantial barrier, as the system requires stable broadband access for cloud-based API functionality. Many Indonesian Puskesmas facilities, particularly in rural areas, lack reliable internet connectivity or face cost constraints that make regular cloud API usage prohibitive. The estimated per-consultation cost of \$0.10-0.15 may appear minimal, but could represent substantial budget impact when scaled across Indonesia's healthcare system. With approximately 10,000 Puskesmas facilities conducting an estimated 100 million consultations annually, the potential API costs could be millions USD per year, a significant consideration for resource-constrained healthcare budgets. Privacy and data governance also represent additional deployment challenges that our proof-of-concept does not address. Current implementation transmits sensitive patient data to external cloud services without encryption or access controls that would be required for HIPAA-equivalent compliance. Indonesian healthcare data protection regulations are still evolving, creating uncertainty about appropriate data handling practices for AI systems processing patient information
\cite{silitonga2024juridical}.

\subsection{Other Limitations and Future Work}

Beyond the infrastructure and clinical limitations discussed above, our experiments revealed several additional technical constraints highlighted during the experiments. Cloud-based architecture creates fundamental limitations for deployment in areas with unreliable connectivity. Rural Puskesmas facilities may lack consistent broadband access, rendering the system unusable during connectivity outages. National data show that around 7\% of Puskesmas have no internet access, and an additional 14\% experience limited or substandard connectivity, especially in remote areas
\cite{aisyah2024internet}.
The system also lacks robust noise reduction or audio quality assessment capabilities. Poor audio quality due to microphone limitations, background noise, or distance from speakers may result in transcription failures or inaccuracies. This represents a critical vulnerability in the busy, acoustically complex environment of typical Puskesmas facilities. Most significantly, current implementation cannot distinguish between doctor and patient speech, potentially leading to misattribution of statements or confusion in clinical summary generation. This limitation could result in serious documentation errors with potential clinical consequences, highlighting the need for speaker diarization capabilities before clinical deployment.

The most critical research priority involves conducting systematic clinical validation in real-world Puskesmas settings across diverse urban, suburban, and rural facilities with varying infrastructure capabilities and patient populations. Consistent with AI-based POMDP clinical models, where accuracy depends on the comprehensiveness of input variables and real-world patient data
\cite{khatim2024toward}, the performance of LLM-driven documentation also hinges on the quality of conversational input and the representativeness of training datasets in Bahasa Indonesia. Both approaches underscore that beyond infrastructure, data quality and contextual adaptation remain central to ensuring safe and effective AI deployment in healthcare
\cite{kim2025context}. 
While our current implementation is able to process a consultation of over five minutes in under 30 seconds, the latency could potentially reduce the anticipated efficiency gains in high-volume clinical settings. This evaluation must include assessment of clinical decision-making impact, patient outcomes, and healthcare provider satisfaction over extended deployment periods, with multiple independent physician reviewers using standardized assessment criteria to establish clinical appropriateness and safety metrics. 

Technical improvements from our current implementation could prioritize offline functionality through edge computing solutions using smaller, locally-deployable models, model distillation techniques for lightweight versions of Whisper and medical summarization systems, and comprehensive security implementations including end-to-end encryption, secure local storage, and audit trail capabilities for compliance monitoring. Additionally, development of Indonesian medical conversation datasets for fine-tuning models specifically for local terminology and consultation patterns will help minimize biases arising from context differences between general-purpose medical datasets and Puskesmas-specific requirements.

\section{Conclusion} \label{conclusion}

The administrative burden of manual documentation in Indonesia's 10,000+ Puskesmas facilities represents a critical barrier to efficient primary healthcare delivery, with providers spending substantial time on data entry rather than direct patient care. In practice, manual recording and reporting can consume up to one-third of consultation time, underscoring the inefficiency of current systems
\cite{juliansyah2024implementation}.
This proof-of-concept study demonstrates the technical feasibility of automated clinical documentation using LLMs to address this challenge that demonstrates succesful integration of Indonesia's ePuskesmas system with an end-to-end framework that transcribes, summarizes, and populates the ePuskesmas form from doctor-patient conversation in under 30 seconds while maintaining clinically appropriate information extraction under the two controlled roleplay scenarios that involve physician validator in Puskesmas Tempilang, Indonesia. The proposed browser-based framework integrates Whisper transcription and GPT-3.5 summarization with existing ePuskesmas workflows underscores a practical approach for resource-constrained healthcare environments where minimal additional changes are required to the existing workflows. Despite the potential, numerous technical limitations and challenges are revealed throughout the experiments, including the need for real-world clinical validation, offline processing capabilities, privacy safeguards, and regulatory compliance.

\section*{Acknowledgment}
The authors are extremely grateful for the system overview and Puskesmas Tempilang management and staff. The authors are also thankful to Bayu Aryo Yudanta (University of Pittsburgh, USA) for sharing state-of-the-arts LLMs. The authors used ChatGPT, Claude, and Gemini for improving grammar and clarity, and have edited and checked the content for factuality.

\section*{Funding Resources}
This work was partially supported by Ikatan Ilmuwan Indonesia Internasional (I-4) Research Grant. 

\bibliographystyle{tfq}
\bibliography{bib}

\appendix
\section{Original Transcript in Indonesian language}

We present the original transcriptions in \autoref{fig:orig_trans_1} and \autoref{fig:orig_trans_1_json} for the first scenario. For the second scenario, the transcriptions are shown in \autoref{fig:orig_trans_2} and \autoref{fig:orig_trans_2_json}.

\begin{figure}[htbp!]
\boxed{
\begin{minipage}{0.96\columnwidth}
\scriptsize
\textit{}
Assalamualaikum Bu, selamat pagi silahkan duduk saya dokter Azmul yang berjaga di puskesmas pada pagi hari ini dengan Ibu siapa? saya Ibu Novi Ibu Novi umurnya berapa Bu? baru 20 sih oh jadi saya panggil Mbak aja ya? iya dok oh iya baik Mbak Novi ada keluhan apa pagi hari ini Mbak? iya dok, perut saya sakit banget terutama di bagian tengah sini dok, yang atas sini oh bagian ulu hati? iya dok ada waktu tertentu gak nyeringnya kapan aja? biasanya sih pagi kalau saya telat sarapan saya kan kuliah ya dok suka masuk pagi jadi sering telat sarapan atau kalau lagi ngerjain tugas malam gitu saya sering telat makan jadi malam-malam sakit dok jadi seringnya itu karena sering telat makan iya dok ada rasa panas gak di tenggorokan yang naik atau sampai lidah terasa pahit? gak sih dok gak ada ya terasa kadang tuh selain perih atau panas perutnya itu juga kayak kembung gitu dok oh ada kembung juga malah, oke ada sampai mual atau muntah Bu? mual sih sering dok, kayak penuh gitu loh dok kayak pengen muntah, tapi gak bisa muntah sih dok akhir-akhir ini lebih terasa cepat kenyang gak kalau makan? iya dok ya makan sedikit itu udah kenyang, gak napsu gitu dok terus kalau habis makan gitu sakitnya itu enakan gak? enakan sih dok, sedikit berkurang dok Mbak Novi seringnya makan apa? sering gak yang kayak pedes-pedes? sering, saya sukanya pedes dok masih sampai sekarang? masih dok, saya kalau gak pedes gak napsu kalau minum kopi atau teh sering gak? seringnya teh dok makan roti? makan roti ya sering juga sarapan seringnya beli roti sih dok oke, oh sering sarapan roti juga ya mungkin sekarang kita periksa dulu ya Mbak mungkin bisa duduk di meja pemeriksaan nanti saya periksa ya, silahkan apakah Mbak Novi berkenan? iya dok, silahkan oke maaf ya Mbak, ini agak diangkat sedikit nanti ini ditemenin sama perawat ya iya dok maaf saya tekan ya, ini nyering gak Mbak? iya dok, yang situ sakit dok nyeri ya nyeri oke, disini? enggak dok disini? enggak dok sini? enggak juga dok oke, ini aja ya? iya aduh dok oh ya oke ya maaf oke, silahkan duduk kembali Mbak saya cuci tangan dulu ya ya, baik Mbak jadi dari keluhan Mbak dan dari hasil pemeriksaan memang ini mengarah pada penyakit MAAG sering disebut MAAG mungkin Mbak juga sudah sering dengar ya oh ya sebelumnya apakah Mbak ada sampai BAB berdarah enggak? berdarah merah dok artinya sekarang lebih ke hitam sih hitam sih enggak dok, luar biasa aja oke, kalau berat badan turun drastis gitu ada enggak? enggak sih dok enggak ada ya, oke baik jadi dari keluhannya dan dari hasil pemeriksaannya memang ini lebih mengarah kepada yang tadi saya katakan, sakit MAAG jadi nanti saya kasih obat dulu untuk seminggu ke depan tolong Mbak diperbaiki dulu nih tidurnya jangan stres-stres juga mungkin Mbak memang lagi kuliah kan tapi usahakan sesekali bisa healing lah ya liburan sama teman-teman, main sama teman-teman kalau weekend gitu dikurangin begadangnya terus makan pedasnya kalau bisa sekarang dikurangin dulu dikurangin, jangan sering-sering terus kopi, tehnya dibatesin dulu gitu ya badannya sih memang untuk beberapa hari ke depan sambil minum obatnya enggak usah dulu gitu ya dok gitu Mbak oke, ada yang mau ditanyakan Mbak? saya tadi minum obatnya berapa lama ya dok? oke, minumnya kita lihat seminggu dulu ya semoga ini bisa membaik ini membaik, kemudian nanti kontrol kasih puskesmas lagi kalau memang ada tanda lebih lanjut artinya mungkin tidak berubah kita perlu evolusi lagi apakah Mbak Novi yang salah minum obat atau memang ternyata gaya hidup Mbak nih yang belum berubah jadi ini PR ya buat Mbak nanti kita lihat seminggu ke depan tolong dijaga makannya yang tadi sudah saya sampaikan dan obatnya diminum sesuai aturan nah nanti saya kasih obat ini ada obat nyiri perutnya sama obat kembungnya nanti diminum semua sebelum makan tapi dikasih jedah nah ini ada nanti obat nanti namanya obat omeprazole diminum 30 menit kalau bisa sebelum makan ada jedahnya 30 menit terus baru antasidanya antasidanya buat ngurangin rasa begahnya juga ya oke baik terima kasih Mbak semoga segera sembuh ya amin, terima kasih banyak dokter
\end{minipage}
}
\caption{Original Version of Gastrointestinal Consultation Transcription}\label{fig:orig_trans_1}
\end{figure}

\begin{figure}[h!]
\vspace{1.6em}
\boxed{
\begin{minipage}{0.96\columnwidth}
\scriptsize
\{\\
\hspace*{1em}``chief\_complaint'': ``Perut sakit terutama di bagian tengah, seringnya saat telat sarapan atau telat makan malam.'',\\
\hspace*{1em}``additional\_complaint'': ``Kembung, sering merasa mual, cepat kenyang, kurang napsu makan.'',\\
\hspace*{1em}``history\_of\_present\_illness'': ``Perut sakit terutama di bagian tengah, seringnya saat telat sarapan atau telat makan malam. Tidak ada rasa panas di tenggorokan atau naik ke lidah. Selain perih atau panas, juga terasa kembung. Sering merasa mual, cepat kenyang, dan kurang napsu makan.'',\\
\hspace*{1em}``past\_medical\_history'': ``Informasi tidak tersedia'',\\
\hspace*{1em}``family\_history'': ``Informasi tidak tersedia'',\\
\hspace*{1em}``recommended\_medication\_therapy'': ``Obat Maag: Omeprazole 30 menit sebelum makan, diikuti antasida untuk mengurangi rasa begah.'',\\
\hspace*{1em}``recommended\_non\_medication\_therapy'': ``Hindari stres, kurangi makan pedas, kurangi begadang, batasi konsumsi kopi dan teh, jaga pola makan.'',\\
\hspace*{1em}``education'': ``Hindari stres, kurangi makan pedas, kurangi begadang, batasi konsumsi kopi dan teh, jaga pola makan. Minum obat Maag, Omeprazole 30 menit sebelum makan, diikuti antasida untuk mengurangi rasa begah.''\\
\}
\end{minipage}
}
\caption{Original Version for Gastrointestinal Consultation JSON}
\label{fig:orig_trans_1_json}
\end{figure}

\begin{figure}
\boxed{
\begin{minipage}{0.96\columnwidth}
\scriptsize
\textit{}
Assalamualaikum, selamat pagi Silahkan duduk Bu Saya dokter Asmul yang menjaga Ibu Sekesmas pagi hari ini ya Udangan Ibu siapa? Iya dok, selamat pagi saya Ibu Desi Oh Ibu Desi Ibu Desi, ada yang bisa saya bantu Bu? Iya dok, ini kepala saya tuh sakit dok Yang sebelah kanan Yang sebelah kanan aja ya Itu nyerinya kayak gimana Bu? Bisa di deskripsikan gak? Maksudnya kayak dia nyut-nyut Atau kayak ditimpa Atau ditekan Atau kayak diikat gitu Berdenyut dok Lebih berdenyut ya Nyut-nyut gitu dok Sampai pelipis sukanya dia Sampai pelipis Ada yang sampai ke mata juga gak? Sampai pelipis Di atas mata aja sih dok Di atas mata aja Di mata enggak ya Enggak sih dok Ya udah berapa lama Ibu ngerasain ini? Sebenernya udah 6 bulanan ini dok Tapi kayak kambungnya 2 minggu terakhir ini dok Kayak gak sembuh-sembuh Itu biasanya Ibu merasa Mulai muncul atau memberatnya itu Pas habis ngapain? Kalau saya tidurnya kurang Misalnya atau stres capek kerja gitu Jadi sekarang Ibu udah tau ya sebenernya apa pencetus-pencetusnya Misalnya kayak kurang tidur gitu Ada stres gitu ya Biasanya tuh 4-8 jaman dok Tapi ini tuh Kalau saya gak kasih minum obat sepanjang hari dok Oh jadi sebelumnya udah coba minum obat? Udah dok tapi beli di warung aja Oh buat warung aja Obat apa kalau Ibu baru tau? Obat paracetamol Paracetamol Biasanya berkurang ya? Dikit dok tapi emang sih Sekarang enggak Ada sampai rasa mual atau muntah? Kalau lagi kambuh ya dok mual Tapi saya gak bisa sampai muntah Ibu itu pas nyerinya itu Ini enggak Kerasa makin berat enggak kalau kayak Habis ngeliat cahaya atau Dengar suara-suara bising gitu Kalau ngeliatan cahaya jadi Iya sih dok kayak bising dikit Kayak makin nambah dikit-dikit ya? Oke Ibu memang pekerjaannya apa? Sampai kurang tidur Saya ini dok jaga di toko dok Jaga di toko Di pasar gitu ya dok Dari sebelum subuh udah bangun Sebelum subuh udah bangun Malamnya tidur jam? Malam juga dok 10.00-11.00 siapin Ibu di rumah ada yang merokok enggak? Ada dok suami saya Berarti Ibu sering Memang maksudnya sering Kecium bau-bau rokok gitu ya? Kalau suami saya sebenarnya Di luar rumah sih dok Tapi kan kalau di pasar banyak ya dok Orang-orang merokok Oke oke baik Ada berat badan yang turun drastis enggak? Enggak sih dok Kalau ini kayak Kerasa kebas Atau lemah Satu sisi badan ada enggak? Enggak sih dok Kebas tuh kesemutan Enggak sih dok Lemah satu badan Kok ini tangan saya susah diangkatnya Sebelah kiri atau sebelah kanan Salah satunya gitu ada enggak? Enggak ada dok Enggak ada ya Oke baik Sekarang mungkin kita Saya mau memastikan dulu Ini memang tidak ada kelainan sarap Iya dok Bisa diikutin ya ikutin jari saya Saya cek matanya dulu ya Coba senyum Bu Julurin lidahnya Lagi Sekarang gembungin pipinya duluan ya Taruh lidahnya di pipi Kanan ya kayak gitu ya Tahan ya Kiri Ibu biasanya ada ini enggak Kalo lagi sakit kepala itu Ada keluar air mata dari mata Atau terus keluar air dari hidung Kadang sih dok Kalo sakit banget Kalo sakit banget Itu kayak pengen nangis aja dok Karena kesakitan aja Tapi kalo biasanya sih enggak Tapi di hidung enggak ada kan Kayak rasanya pengen nangis Oke baik baik Oke ibu jadi dari Perasanan amnesis Saya tanya jawab dengan ibu Dan hasil pemeriksaannya Untuk saat ini Ibu bisa dikatakan Terkena migrain Jadi itu memang khas sakit kepala Sebelah aja dan Berdenyut dan sejauh ini Memang ini lebih Karena Ibu kurang tidur Terus aktivitasnya mungkin berat Ada stresor-stresor yang lain Saran saja ini banyak yang minum air putih Minum air putih Terus istirahatnya yang cukup Nanti juga saya kasih obat Ini pada prinsipnya obatnya Hanya mengendalikan Karena ini memang bisa dikatakan sakit kepala Yang pencetusnya ini Tidak khas Jadi cukup diubah pola hidupnya Terus sambil minum obat Yang mengendalikan sakit kepalanya Insya Allah nanti bisa Lebih baik lagi Ada yang mau ditanyakan bu? Ini parah sampe perlu CT scan gitu Gitu gak dong di kepala saya Oke baik saya mengerti Kekhawatiran ibu jadi memang Sakit kepala seperti ini kita juga Harus waspada adanya Kelainan di dalam kepala Nah salah satu tanda Untuk menyingkirkannya itu terlebih dahulu Kita memang memeriksa sarap-sarap ibu Makanya tadi saya minta tolong gerakin mata Keluar dari lidah Pengen ngeliat kekuatan pipi Terus nanya apakah ada kelemahan Dan saat ini memang tidak ada Kelainan untuk pemeriksaan fisiknya Jadi saat ini kita tidak perlu dulu Melakukan pemeriksaan CT scan Seperti itu bu Jadi ini saya obatin dulu Sama hindarin yang dokter bilang tadi Ya kita lakukan itu dulu aja Nanti kontrol aja Satu minggu ke depan ya bu Oke dok Ada lagi bu yang mau tanya Udah dok Terima kasih ya bu Mudah-mudahan cepat sembuh juga ya Terima kasih dok
\end{minipage}
}
\caption{Original Version of Migraine Consultation Transcription}
\label{fig:orig_trans_2}
\end{figure}

\begin{figure}
\boxed{
\begin{minipage}{0.96\columnwidth}
\scriptsize
\{\\
\hspace*{1em}``chief\_complaint'': ``Sakit kepala sebelah kanan selama 6 bulan, kambuh 2 minggu terakhir, berdenyut hingga pelipis dan atas mata.'',\\
\hspace*{1em}``additional\_complaint'': ``Tidak ada informasi tambahan.'',\\
\hspace*{1em}``history\_of\_present\_illness'': ``Sakit kepala sebelah kanan berdenyut selama 6 bulan, kambuh 2 minggu terakhir, terutama setelah kurang tidur dan stres. Nyeri seperti ditimpa, berdenyut hingga pelipis dan atas mata, disertai mual saat kambuh.'',\\
\hspace*{1em}``past\_medical\_history'': ``Informasi tidak tersedia.'',\\
\hspace*{1em}``family\_history'': ``Informasi tidak tersedia.'',\\
\hspace*{1em}``recommended\_medication\_therapy'': ``Minum obat paracetamol untuk mengendalikan sakit kepala.'',\\
\hspace*{1em}``recommended\_non\_medication\_therapy'': ``Banyak minum air putih, istirahat cukup, ubah pola hidup dengan menghindari kurang tidur dan stres.'',\\
\hspace*{1em}``education'': ``Migrain sebelah kanan, disarankan minum obat paracetamol, minum air putih, istirahat cukup, hindari kurang tidur dan stres. Tidak perlu CT scan saat ini.''\\
\}
\end{minipage}
}
\caption{Original Version for Migraine Consultation JSON}
\label{fig:orig_trans_2_json}
\end{figure}

\end{document}